# Empirical Evaluation of Leveraging Named Entities for Arabic Sentiment Analysis

Hala Mulki[1], Hatem Haddad[2], Mourad Gridach[3] and Ismail Babaoğlu[1]

[1]Computer Engineering Department, Konya Technical University, Turkey
[2]RIADI Laboratory, National School of Computer Sciences, University of Manouba, Tunisia
[3]Computational Bioscience Program, University of Colorado, School of Medicine, USA

**Abstract:** *Social media reflects the public's attitudes towards specific events. Events are often related to persons, locations or organizations, the so-called Named Entities (NEs). This can define NEs as sentiment-bearing components. In this paper, we dive beyond NEs recognition to the exploitation of sentiment-annotated NEs in Arabic sentiment analysis. Therefore, we develop an algorithm to detect the sentiment of NEs based on the majority of attitudes towards them. This enabled tagging NEs with proper tags and, thus, including them in a sentiment analysis framework of two models: supervised and lexicon-based. Both models were applied on datasets of multi-dialectal content. The results revealed that NEs have no considerable impact on the supervised model, while employing NEs in the lexicon-based model improved the classification performance and outperformed most of the baseline systems.*

## 1. Introduction

Social media represents a huge source of information from which opinions can be extracted and exploited in many analytical studies. Sentiment Analysis (SA) is a Natural Language Processing (NLP) task that mines the subjective content in a piece of text and categorizes it into positive, negative or neutral polarity using computational linguistics techniques [17]. SA can be performed at three levels of granularity: (a) Document-level: where a piece of text is analyzed as a whole to produce an overall sentiment, (b) Entity-level: recognizes the sentiment of specific aspects in a piece of text and (c) Sentence-level: provides the sentiment for each sentence in the corpus. Recently, most studies focused on sentence-level SA as the opinions on social media are mostly shared in the form of sentences.

Social media often combines the opinions of the public towards all NE types (persons, locations or organizations). Thus, NEs in a sentence can be considered essential components without which the subjectivity of the sentence might be lost. To clarify that, in "من لم ينتخب نداء تونس كأنه انتخب حركة النهضة"[1], there are two NEs: "حركة النهضة"[3] and "نداء تونس"[2]; if we omit these NEs, the subjectivity of the sentence cannot be recognized while with them retained, the tweet's polarity would not be correctly inferred unless the sentiment borne by each NE is identified. In addition, the polarity of a tweet, containing an NE and posted during a specific period of time, is affected by this very NE and the attitudes towards it at that time. For example, when exploring the dataset of [4] that was collected during the post-revolution Tunisian elections, we find that 80% of the tweets that contained the NE "بن علي", which refers to the former Tunisian president, has a negative sentiment. Similarly, in the dataset of [1], the location name "سوريا" meaning "Syria" that has been recently related to war incidents, was encountered in 30 tweets, 75% of them was negative.

Therefore, given a Twitter/Facebook dataset collected in a certain period of time, we hypothesize that recognizing the sentiment of an NE can contribute in identifying the polarity of the sentence in which it is mentioned. NEs sentiment recognition is not trivial and has not been tackled in previous studies; as most of them focused on NEs recognition rather than NEs exploitation for further NLP tasks. Moreover, combining Named Entity Recognition (NER) with sentence-level SA poses another level of difficulty especially when Arabic language is tackled. On one hand, compared to Indo-European languages, Arabic texts has no notion of capital letters, therefore, Arabic NER systems have to recognize NEs without using capitalization among the features. On the other hand, Arabic sentiment analysis (ASA) is challenging especially with the existence of two Arabic language variants: Modern Standard Arabic (MSA) and Dialectal Arabic (DA) where the latter is commonly used in social media. Both variants have a complex morphology as words are of a highly inflectional and derivational nature [19] such that some adjectives and NEs might be identical. This is because many Arabic person names are derived from adjectives as in the positive adjective "سعيد" that means ``happy'' and can be also used as a male person name which misleads the sentiment classifier. To avoid such confusion, NEs are usually recognized then person names are eliminated while mining the sentiment [13,19]. Finally, most of the Arabic NLP resources needed for NER or SA are not publicly available.

---

[1] Those who didn't vote for Nidaa Tounes, as if they voted for Ennahda Movement.
[2] Nidaa Tounes Party.
[3] Ennahda Movement Party.

Here, we present an empirical evaluation of the effectiveness of NEs in inferring the sentiment of MSA/DA tweets and Facebook comments. To the extent of our knowledge, this is the first effort to pair NEs with ASA. While previous ASA works ignored NEs or eliminated some NE types, we investigate using NEs as expressive features to be included in ASA framework of two model variants: supervised and lexicon-based. This is done by classifying the NEs extracted via the NER system into positive or negative. The sentiment-annotated NEs are then replaced with special tags in the corpus. The proposed framework was applied on four datasets of MSA/DA content. We conducted SA once with NEs tagged and included among the features then with them considered as ordinary tokens. This enabled answering these research questions:

- What is the impact of including NEs on ASA models: lexicon-based and supervised?
- For datasets rich of NEs, is it more likely to have a better SA performance?
- Are NEs reliable enough to infer the Arabic sentiment? For which SA models?

## 2. Arabic Sentiment Analysis

Arabic SA methods can be categorized under two main categories: machine learning and lexicon-based.

### 2.1. Machine Learning Methods

These methods adopt supervised/unsupervised learning strategies using either hand-crafted features or distributed text representations. The training process depends on learning that a combination of specific features yields a certain polarity [17].

Among the ASA systems that employed hand-crafted-features, we can refer to [14] where bag-of-words along with several levels of stemming were used to train a supervised sentiment classifier of MSA/Jordanian tweets. The best algorithm was SVM with an accuracy of 87.2%.

In the same context, the authors in [4] presented a supervised SA system for Tunisian tweets. With different bag-of-word schemes used as features, binary and multiclass classifications were conducted. SVM was found of the best results for binary classification with an accuracy of 71.09% and an F-measure of 63%.

A novel SA model based on text embeddings was proposed by [1]. The model was trained with Arabic word embeddings generated via word2vec [21] and applied on MSA/DA datasets. Among the used classification algorithms, Nu-SVM scored the best results with an accuracy of 80.21% and an F-measure of 79.62% for the twitter dataset.

In [18], doc2vec algorithm [16] was used to produce document embeddings of Tunisian comments. The generated embeddings were fed to train a multi-layer perceptron (MLP) classifier where the achieved accuracy and F-measure values were both 78%.

### 2.2. Lexicon-based Methods

The core components of such models are manually-built, predefined or translated sentiment lexicons. A sentiment lexicon contains subjective words along with their polarities (positive or negative) and polarity scores also known as weights [17]. Thus, the polarity of a word or a sentence can be determined by one of the following algorithms:

- Straight Forward Sum (SFS): adopts the uniform weighting scheme, where negative words have the weight of -1 while positive ones have the weight of 1. The polarity of a given text is calculated by accumulating the weights of negative and positive terms. The sign of the resulting sum infers a positive sentiment if it is positive and a negative one if it is negative [19].

- Double Polarity (DP): Assigns both a positive and a negative weight for each entry in the lexicon. For example, if a positive term has a score of 0.7, then its negative score: - (1+0.7) = -0.3 and vice versa for negative terms. To define the polarity of a sentence, two scores are accumulated: the positive and negative where the one of the greater absolute value defines the total sentiment [19].

In [15], manually-built lexicons were compared against automatically-built ones for the SA task. Each lexicon type was expanded by adding synonyms, stemming and most common words via Term Frequency (TF) weighting, emotions and dialectal terms. Datasets from [8] and [14] were used in the experiments. The study showed that stemming degraded the performance for both lexicon types. The merged lexicon with light stemming achieved the best accuracy equals to 74.6%.

In [12], a lexicon-based SA system was used to classify the sentiment of Tunisian tweets. The author developed a Tunisian morphological analyzer to produce linguistic features. Using a Tunisian version, the model achieved an accuracy of 72.1% considering only the positive/negative tweets.

## 3. Arabic Named Entity Recognition

NER approaches fall into three categories: rule-based, Machine Learning (ML) and hybrid of both. For the English language, the state-of-the-art research is dominated by ML approaches. Systems that employ deep neural networks as the main building blocks have recently become the dominant methods among ML approaches. These systems combine various Neural Network (NN) architectures with other ML methods to handle NEs. They rely on recurrent neural networks (RNNs), convolutional neural networks (CNNs) and conditional random fields (CRF). In addition, NN-based approaches do not rely on hand-crafted features as they use pre-trained word embeddings to initialize the word vectors and character-level embeddings [2].

For the Arabic language, [10] developed an Arabic NER for social media based on deep neural networks. This system achieved the state-of-the-art on the Twitter dataset of [5] where it scored an F-measure of 85.71%. In our study, we employed this system to recognize Arabic NEs in the input data.

## 4. Methodology

In this study, SA of tweets and Facebook comments was conducted via a framework called Tw-StAR with supervised and lexicon-based model variants. NEs extraction was performed by the system of [10]. The extracted NEs were then fed to our NEs sentiment detection algorithm with which the polarity of an NE is identified. Later, the sentiment-annotated NEs were included in Tw-StAR framework to assist in the sentiment classification task. For both model variants, we adopted n-gram schemes such that they were fed to train the classifiers of the supervised model, while they were looked up in the sentiment lexicon adopted by the lexicon-based model. The general pipeline of the proposed framework is illustrated in Figure 1.

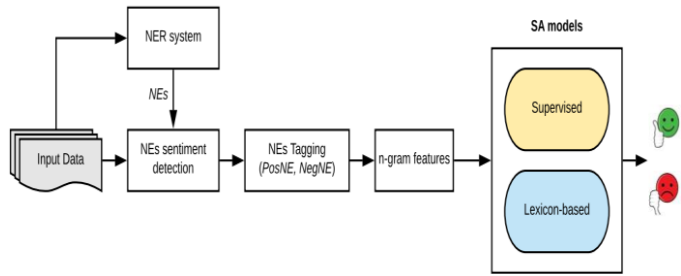

Figure 1. Tw-StAR sentiment analysis framework.

### 4.1. Arabic NER system

To extract Arabic NEs, we used the system of [10]. This system can handle NEs encountered in social media data. Figure 2 shows its main architecture while recognizing the NEs of the sentence " ريال مدريد يفوز بالدوري الاسباني"[4].

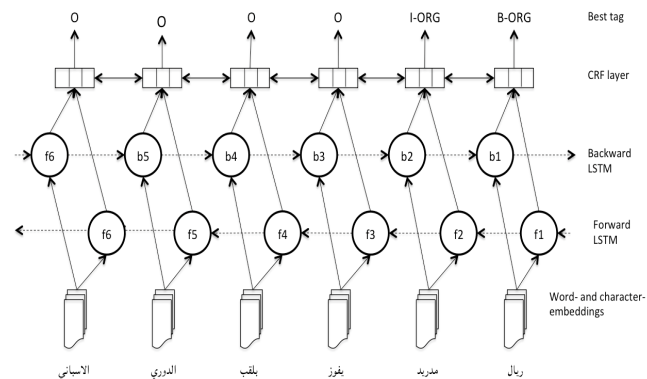

Figure 2. The architecture of the Arabic NER system.

---

[4] Real Madrid wins the Spanish league title.

The input is a set of word vectors obtained using the following strategy: Each word in the sentence is composed by the concatenation of two vectors, the first is resulted from a lookup table containing the pre-trained word embeddings while the second was created by character-level embeddings. The model then uses a bidirectional Long Short Term Memory (LSTM) networks to read the sentence in two directions (we note that Arabic texts are written from right to left) where each word receives left and right representations. Both representations are then concatenated and linearly projected onto the next layer.

It should be noted that each LSTM direction has its own parameters. Finally, a CRF layer is used on the top of the bidirectional LSTM in order to capture contextual features in the form of neighboring named entity recognition tags. To train the model, Backpropagation Through Time (BBTT) algorithm was used to update parameters together with Stochastic Gradient Descent (SGD) with fixed learning rate. The dimension of word embeddings is 100 with 25 for each character.

### 4.2. NEs Sentiment Detection Algorithm

To include NEs in the SA task and inspired by SFS method [19], we have developed the algorithm illustrated in Figure 3 to detect the sentiment of the NEs extracted by the system of [10].

```
Algorithm 1: NEs sentiment detection Algorithm
  Data: tweet tokens: T, Named Entity: N, tweet polarity: pol_t, NE computed
        score: N_score
  Result: NE assigned polarity: N_pol
  N_score ← 0
  foreach N in NEs do
      foreach tweet in corpus do
          if N ⊂ T then
              if pol_t=positive then
                  increase N_score by 1
              else if pol_t=negative then
                  decrease N_score by 1
          else
              increase N_score by 0
      end
  end
  if N_score>0 then
      N_pol = positive
  else if N_score<0 then
      N_pol = negative
```

Figure 3. NEs sentiment detection algorithm.

According to our algorithm, the polarity of an NE in a corpus is defined by the majority of attitudes towards it. In other words, the sentiment of each NE in the dataset is identified as positive or negative according to how frequently this NE is mentioned within positive or negative tweets. This can address the confusion of detecting the sentiment of two NEs that have contradict polarities and mentioned in the same tweet as in " إنه خالد بن الوليد القائد النبيل يا من يفتخر بـ خالد بن "[5] where "هتلر"[6] known as a dictator and " هتلر

---

[5] To those who boast of Hitler, it is Khalid Ibn Al-Walid, the noble leader.

[6] Hitler.

"الوليد"[7], who was a commander, were mentioned together in a positive tweet. In this case, the algorithm gives both NEs a positive score at the beginning, however, after browsing the rest of tweets the score of "هتلر" will decrease since it is mostly mentioned in negative contexts while the score related to " خالد بن الوليد" will increase if the majority of the tweets containing it are positive.

### 4.3. Sentiment Analysis using Tw-StAR

The general SA pipeline adopted for both models can be briefly described as the following:

- Preprocessing: the input data was normalized such that URLs, tweet-related symbols, punctuation and non-Arabic characters were removed, while Stopwords and negations were retained to enable capturing the sentiment borne by compound terms. So, the tweet: "فيه من ريحة الغالي! #هاري_بوتر :)" becomes after preprocessing: "فيه من ريحة الغالي هاري بوتر".
- NEs tagging: after NEs are extracted from the training corpus and their polarities are detected, every NE is replaced with either a positive tag (PosNE) or a negative one (NegNE) in both training and test divisions of each dataset.
- Feature extraction: several n-gram schemes were generated as features for both models. In the supervised model, unigrams, bigrams, trigrams and combinations of them were adopted as they can capture information about the local word order and save the training time. However, in the lexicon-based model, unigrams and a combination of unigrams and bigrams were used to cover single and compound phrases of the used lexicon [20].
- Sentiment classification: supervised algorithms were employed by the supervised model, while DA lexicons along with the scoring algorithm SFS [19] were used to detect the sentiment in the lexicon-based model.

### 4.4. Tw-StAR supervised Model

In this model, the feature vector of each sentence is constructed via examining the presence/absence of the n-gram schemes among the sentence's tokens. The produced n-gram schemes include trigrams in addition to unigrams and bigrams since higher-order n-grams can better capture the contextual information [7]. Later, feature selection was conducted using the Term Frequency (TF) weighting by FreqDist module. The supervised model was trained using NB from Scikit-Learn and linear SVM from LIBSVM.

Having all the NEs recognized, identified as having positive or negative polarity and tagged properly, they were involved in inferring the sentiment. This is because the tags of NEs are included among the tweets' n-grams from which the feature vectors are constructed. Figure 4 shows the pipeline of this model.

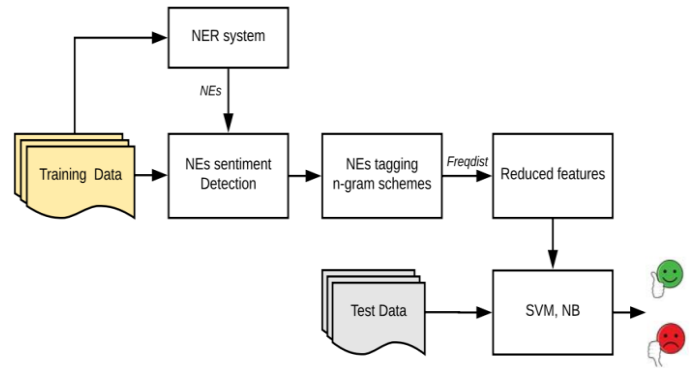

Figure 4. Tw-StAR supervised model with NEs included.

### 4.5. Tw-StAR Lexicon-based Model

This model uses an integrated lexicon constructed out of pre-built lexicons: MSA/Egyptian NileULex [20], MSA/DA seeds from Arabic Emotion Lexicon (AEL) and Arabic Hashtag Lexicon seeds (AHL) [6,7] plus two manually-built lexicons for Levantine (LevLex) and Gulf (GulfLex) dialects. For the Tunisian datasets, we built a Tunisian lexicon (TunLex). Table 1 lists these lexicons and their sizes.

Table 1. The used sentiment lexicons.

| Sentiment Lexicon | Positive | Negative | Size |
|---|---|---|---|
| NileULex | 1697 | 4256 | 5953 |
| AEL | 12 | 11 | 23 |
| AHL | 107 | 118 | 225 |
| LevLex | 258 | 559 | 817 |
| GulfLex | 33 | 67 | 100 |
| TunLex | 1953 | 3329 | 5282 |

To recognize the sentiment of the input data, the tokens of a sentence either unigrams or the combination of unigrams and bigrams are looked up in the proper lexicon. When a match is found, the sentence's polarity score is calculated using the SFS algorithm. Similar to the supervised model, the tags of NEs were included in the n-gram features to be looked up in the lexicon. Consequently, both NEs tags: PosNE and NegNE were added to the lexicon as positive and negative entries having the scores of 1 and 0, respectively. Fig 5 shows the pipeline of the lexicon-based model.

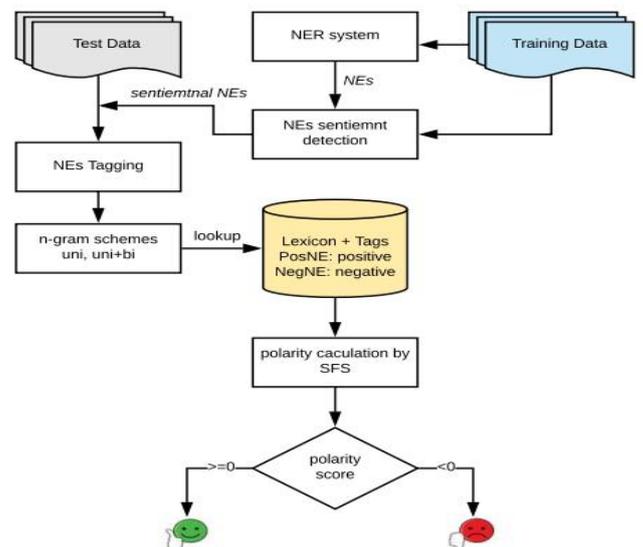

Figure 5. Tw-StAR lexicon-based model with NEs included.

---
[7] Khalid Ibn Al-Waleed

## 5. Experiments and Results Evaluation

### 5.1. Datasets:

The proposed system was applied on these datasets:

- Jordanian Egyptian Gulf (JEG): was tackled in [3], it combines positive/negative tweets of the dialects: (a) Jordanian: ArTwitter [14], (b) Egyptian: ASTD [11] and (c) Gulf: QCRI [9].
- Tunisian Arabic Corpus (TAC): consists of tweets about media, telecom and politics. It was collected by [12] and annotated for positive, negative and neutral polarity.
- Tunisian Election Corpus (TEC): refers to a set of MSA/Tunisian tweets collected by [4] during the Tunisian elections in 2014.
- Tunisian Sentiment Analysis Corpus (TSAC): collected by [18], it combines positive and negative Facebook comments about TV shows.

Each dataset was divided into training and test sets as it is shown in Table 2.

Table 2. Polarity distribution in the training and test sets.

| Dataset | Training | | Test | | Total size |
|---|---|---|---|---|---|
| | Positive | Negative | Positive | Negative | |
| JEG | 1732 | 1702 | 415 | 445 | 4294 |
| TAC | 306 | 290 | 76 | 74 | 746 |
| TEC | 968 | 1466 | 276 | 333 | 3043 |
| TSAC | 2782 | 3451 | 672 | 890 | 7795 |

### 5.2. Named Entities Results

The statistics of the extracted NEs are listed in Table 3 as E-NEs, Pos-NEs, Neg-NEs and A-NEs denote the number of the extracted NEs, positive NEs, negative NEs and the sentiment-annotated NEs, respectively.

Table 3. NEs statistics extracted from each dataset.

| Dataset | E-NEs | Pos-NEs | Neg-NEs | A-NEs |
|---|---|---|---|---|
| JEG | 841 | 395 | 487 | 782 |
| TAC | 240 | 99 | 129 | 228 |
| TEC | 658 | 192 | 410 | 602 |
| TSAC | 615 | 198 | 350 | 548 |

We notice that from large-sized datasets such as TSAC, JEG and TEC, more NEs could be extracted. In addition, in all datasets, the number of negative NEs is greater than that of the positive ones.

On the other hand, although the Tunisian datasets: TSAC and TEC have a larger or close size compared to JEG; yet the used NER system extracted less NEs from them compared to those extracted from JEG. This is due to the fact that the used NER system employed pre-trained word embeddings from [5] produced with corpora composed of MSA, Egyptian and Levantine content which is quite similar to that of JEG while it is far from the Tunisian dialect. This made most of the Tunisian terms unknown when they looked up in the lookup table of the NER system; therefore they will be initialized randomly instead of being initialized with pre-trained word embeddings.

### 5.3. Sentiment Analysis Results

#### 5.3.1. Supervised Model Results

The supervised model was trained once without tagging NEs (Tw-StAR) then with NEs tagged and included in the features (Tw-StAR+NEs). Three experiment variants were conducted, where the first involved using all n-gram features, while the second and third used a reduced number of features obtained by the TF scheme for the thresholds: 2 and 3, respectively.

We chose to review the results of the experiment of the best averaged F-measure, with/without NEs. Table 4 lists this model's results where uni, bi and tri refer to unigrams, bigrams and trigrams, respectively. While Prec, Rec, F1 and Acc indicate the averaged precision, recall, F-measure and accuracy, respectively. A comparison with baseline systems is shown in Table 5.

Table 4. Supervised Tw-StAR performance for all datasets.

| Dataset | NEs | Algorithm | Prec (%) | Rec (%) | F1 (%) | Acc (%) |
|---|---|---|---|---|---|---|
| JEG | No | NB | **77.0** | **77.0** | **76.9** | **76.9** |
| | | SVM | 71.6 | 71.2 | 71.2 | 71.4 |
| | Yes | NB | **76.8** | **76.8** | **76.7** | **76.7** |
| | | SVM | 69.9 | 69.8 | 69.8 | 69.9 |
| TAC | No | NB | 83.4 | 81.9 | 81.8 | 82.0 |
| | | SVM | **85.2** | **84.6** | **84.6** | **84.7** |
| | Yes | NB | **84.4** | 83.2 | 83.2 | 83.3 |
| | | SVM | 83.4 | **83.3** | **83.3** | **83.3** |
| TEC | No | NB | 71.8 | 68.8 | 68.7 | 70.4 |
| | | SVM | **75.0** | **71.4** | **71.4** | **73.1** |
| | Yes | NB | 72.3 | 69.6 | 69.6 | 71.1 |
| | | SVM | **74.4** | **71.2** | **71.2** | **72.7** |
| TSAC | No | NB | 91.2 | 92.0 | 91.4 | 91.4 |
| | | SVM | **92.8** | **92.5** | **92.7** | **92.8** |
| | Yes | NB | 91.6 | **92.4** | 91.7 | 91.7 |
| | | SVM | **92.4** | 92.2 | **92.3** | **92.4** |

Table 5. Supervised Tw-StAR against baselines.

| Dataset | Model | Prec (%) | Rec (%) | F1 (%) | Acc (%) |
|---|---|---|---|---|---|
| JEG | word2vec+supervised [3] | 76.5 | **83.0** | **79.6** | **80.2** |
| | Tw-StAR | **77.0** | 77.0 | 76.9 | 76.9 |
| | Tw-StAR + NEs | 76.8 | 76.8 | 76.7 | 76.7 |
| TAC | lexicon-based [12] | 63.0 | 72.9 | 67.3 | 72.1 |
| | Tw-StAR | **85.2** | **84.6** | **84.6** | **84.7** |
| | Tw-StAR + NEs | 83.4 | 83.3 | 83.3 | 83.3 |
| TEC | supervised + n-grams [2] | 67.0 | 71.0 | 63.0 | 71.1 |
| | Tw-StAR | **75.0** | **71.4** | **71.4** | **73.1** |
| | Tw-StAR + NEs | 74.4 | 71.2 | 71.2 | 72.7 |
| TSAC | Doc2vec + MLP [8] | 78.0 | 78.0 | 78.0 | 78.0 |
| | Tw-StAR | **92.8** | **92.5** | **92.7** | **92.8** |
| | Tw-StAR + NEs | 92.4 | 92.2 | 92.3 | 92.4 |

#### 5.3.2. Lexicon-based Model Results

In this model, Tw-StAR experiments were conducted considering two lexicons: (a) an integrated lexicon constructed out of NileULex, AEL, AHL, LevLex and GulfLex (see Table 1) to handle JEG dataset whose content combines Egyptian, Levantine and gulf dialects in addition to MSA and (b) TunLex to mine the sentiment of TAC, TEC and TSAC datasets. In Tw-StAR+NEs experiments, however, the same previous lexicons were used but with positive and negative NEs tags: PosNE, NegNE added as entries having positive and negative scores, respectively. The sentiment detection procedure was carried out by looking for a tweet's unigrams (uni) then unigrams and bigrams (uni+bi) in the relevant lexicon, once with NEs tagged then with them treated as ordinary tokens.

Table 6. Lexicon-based Tw-StAR performance for all datasets.

| Dataset | NEs | Features | Prec (%) | Rec (%) | F1 (%) | Acc (%) |
|---|---|---|---|---|---|---|
| JEG | No | uni+bi | **71.6** | 67.4 | 66.3 | 68.3 |
| | Yes | uni+bi | 70.7 | **69.2** | **68.9** | **69.7** |
| TAC | No | uni+bi | 66.9 | 66.7 | 66.6 | 66.7 |
| | Yes | uni+bi | **70.8** | **70.6** | **70.6** | **70.7** |
| TEC | No | uni+bi | 66.6 | 61.5 | 59.8 | 64.0 |
| | Yes | uni+bi | **69.1** | **65.6** | **65.0** | **67.5** |
| TSAC | No | uni+bi | 84.5 | 83.8 | 81.8 | 81.8 |
| | Yes | uni+bi | **84.6** | **84.7** | **82.8** | **82.8** |

The best results of the lexicon-based are shown in Table 6. These performances were then compared against baseline systems as it is shown in Table 7.

Table 7. Lexicon-based Tw-StAR performance against baselines

| Dataset | Model | Prec (%) | Rec (%) | F1 (%) | Acc (%) |
|---|---|---|---|---|---|
| JEG | word2vec+supervised [3] | 76.5 | 83.0 | 79.6 | 80.2 |
| | Tw-StAR | 71.6 | 67.4 | 66.3 | 68.3 |
| | Tw-StAR + NEs | 70.7 | 69.2 | 68.9 | 69.7 |
| TAC | lexicon-based [12] | 63.0 | **72.9** | 67.3 | **72.1** |
| | Tw-StAR | 66.9 | 66.7 | 66.6 | 66.7 |
| | Tw-StAR + NEs | **70.8** | 70.6 | **70.6** | 70.7 |
| TEC | supervised + n-grams [2] | 67.0 | **71.0** | 63.0 | **71.1** |
| | Tw-StAR | 66.6 | 61.5 | 59.8 | 64.0 |
| | Tw-StAR + NEs | **69.1** | 65.6 | **65.0** | 67.5 |
| TSAC | Doc2vec + MLP [8] | 78.0 | 78.0 | 78.0 | 78.0 |
| | Tw-StAR | 84.5 | 83.8 | 81.8 | 81.8 |
| | Tw-StAR + NEs | **84.6** | **84.7** | **82.8** | **82.8** |

## 6. Results Discussion

When exploring the performances of the supervised model in Table 4, we notice that although tagging NEs in the training corpus was expected to enhance the performance as it decreases the features' size by reducing all the NEs to either PosNE or NegNE terms, a degraded performance could be noticed. While comparable results were scored with and without NEs in JEG, TEC and TSAC datasets, the performance degraded when NEs were added in TAC as the F-measure decreased by 1.3%. This could be due to the fact that inferring the sentiment using n-gram schemes depends on capturing the contextual information with which a specific n-gram scheme is learned to be an indicator of a specific sentiment. As the sentiment of an NE was deduced based only on how frequent it is mentioned within a context of a positive or a negative polarity regardless of the context's words, it is possible for a positive NE to be included within a negative context (n-gram scheme) and vice versa which misleads the classifier.

Unlike the supervised model, the performance of the lexicon-based model was favorably impacted by involving NEs in the SA task. As it can be seen in Table 6, for uni+bi features, the sentiment classification performance with NEs considered and NE tags added to the lexicons could outperform the one obtained by the ordinary lexicons. Indeed, the evaluation measures increased in all datasets as the F-measure values of Tw-StAR+NEs were 68.9%, 70.6%, 65% and 82.8% compared to 66.3%, 66.6%, 59.8% and 81.8% achieved by Tw-StAR for JEG, TAC, TEC and TSAC datasets, respectively. The reason behind such improvement is that uniform weight scheme lexicons ignore the contextual-related information where a sentence's polarity is defined based on the polarity scores of its constituent words [13,19]. This in turn enables the sentiment-annotated NEs deduced regardless of the context, to effectively contribute in recognizing the polarity of the tweet containing it. Moreover, with NEs tagged in the test corpus, it became possible to employ NEs of the type person names in the SA task. Hence, the issue caused by confusing a person name with an adjective could be avoided without the need to eliminate person names as in [13,19].

Considering Table 7 which compares the lexicon-based model against the baseline systems, it should be noted that this comparison is meaningful only for TAC dataset where the baseline system [12] is a lexicon-based one; though we observed that Tw-StAR+NEs outperformed the baselines in Tunisian datasets: TAC, TEC and TSAC. This could be explained by the positive impact of NEs on the polarity detection in addition to the good coverage provided by the used Tunisian lexicon. In contrast, it is reasonable that the performance degraded in JEG dataset as the F-measure decreased by 10.7% compared to [1] that used pre-trained word embeddings. In addition, the efficiency of Tw-StAR with and without NEs can be attributed to the looking up for uni+bi tokens in the lexicon increases the matching ratios of compound terms.

Finally, for datasets rich of NEs (see Table 3); we could not determine the impact of the number of the sentiment-annotated NEs on SA within Tw-StAR+NE lexicon-based model. To clarify that, although JEG has the greatest number of sentiment-annotated NEs, the improvement recorded in the F-measure value was 2.6%, while for TEC that has less annotated NEs, the F-measure increased by 5.2%. We believe that the performance of the lexicon-based model for a specific corpus, with NEs included, is not related to the number of the sentiment-annotated NEs in the corpus as much as it is to the consistency of that corpus. More specifically, in a corpus having a good degree of consistency, the training and test data tend to contain more similar NEs. Thus, it is more likely to have a consensus on the sentiment of a specific NE which leads to an accurate sentiment assignment of that NE and hence to a better sentiment classification.

## 7. Conclusion and Future Work

We presented a pioneering step towards leveraging NEs in Arabic sentiment analysis. It was observed that NEs can form reliable indicators of Arabic sentiment if they are included within the lexicon-based model of Tw-StAR framework, while a similar behavior could not be noticed in the supervised model. In addition, adding NEs to the lexicon-based model enabled the exploitation of person names in the SA task. On the other hand, it was revealed that the impact of the number of sentiment-annotated NEs is less important than the consistency of the tackled corpora as the latter affects the sentiment classification performance. For the future work, we will investigate the impact of NEs on multiclass SA. Furthermore, as NEs performed better in the lexicon-based model, it would be interesting to develop a sentiment recognition method that considers the negation and sarcasm. Regarding the NER task, it could be enhanced if the different writing styles of NEs are handled. Lastly, Tw-StAR framework would be further examined on datasets of other languages such as English, French and Turkish.

## References

[1] A. Aziz A., and Lixin T., "Word embeddings for Arabic sentiment analysis", *IEEE International Conference on Big Data (Big Data)*, pp. 3820-3825, 2016.

[2] Chen D., and Christopher M., "A fast and accurate dependency parser using neural networks", *Proceedings of 2014 conference on empirical methods in natural language processing (EMNLP)*, pp. 740-750.

[3] Kareem D., "Named entity recognition using cross-lingual resources: Arabic as an example", *Proceedings of the 51st Annual Meeting of the Association for Computational Linguistics (Volume 1: Long Papers)*, vol. 1, pp. 1558-1567, 2013.

[4] Karim S. et al., "Tunisian Dialect and Modern Standard Arabic Dataset for Sentiment Analysis: Tunisian Election Context", *To appear in IEEE proceedings of ACLing 2016*.

[5] Mohamed Z., et al., "Word representations in vector space and their applications for Arabic", *International Conference on Intelligent Text Processing and Computational Linguistics*, pp. 430-443, 2015.

[6] Mohammad S., Saif M., and Svetlana K., "Sentiment after translation: A case-study on Arabic social media posts", *Proceedings of the 2015 conference of the North American chapter of the association for computational linguistics: Human language technologies*, pp. 767-777, 2015.

[7] Mohammad S., Saif M., and Svetlana K., "How translation alters sentiment", *Journal of Artificial Intelligence Research*, vol. 55, pp. 95-130, 2016.

[8] Mohammed A., Nawaf A., and Mahmoud A., "An analytical study of Arabic sentiments: Maktoob case study", *8th International Conference for Internet Technology and Secured Transactions (ICITST)*, pp. 89-94, 2013.

[9] Mourad A., and Karim D., "Subjectivity and sentiment analysis of modern standard Arabic and Arabic microblogs", *Proceedings of the 4th workshop on computational approaches to subjectivity, sentiment and social media analysis*, pp. 55-64. 2013.

[10] Mourad G., "Character-Aware Neural Networks for Arabic Named Entity Recognition for Social Media", *Proceedings of the 6th Workshop on South and Southeast Asian Natural Language Processing (WSSANLP)*, pp. 23-32, 2016.

[11] Nabil M., Mohamed A., and Amir A., "Astd: Arabic sentiment tweets dataset", *Proceedings of the 2015 Conference on Empirical Methods in Natural Language Processing (EMNLP)*, pp. 2515-2519, 2015.

[12] Nadia K., *Tunisian Arabic Customer's Reviews Processing And Analysis For an Internet Supervision System*, Sfax University, 2017.

[13] Najwa E., et al., "Sentiment analysis of colloquial Arabic tweets", *ASE BigData, Social Informatics, PASSAT, BioMedCom Conference*, Harvard University, pp. 1-9, 2013.

[14] Nawaf A., et al., "Arabic sentiment analysis: Lexicon-based and corpus-based", *IEEE Jordan Conference on Applied Electrical Engineering and Computing Technologies (AEECT)*, pp. 1-6, 2013.

[15] Nawaf A., et al., "*Towards improving the lexicon-based approach for arabic sentiment analysis*", *International Journal of Information Technology and Web Engineering (IJITWE)*, vol.9, no.3, pp. 55-71, 2014.

[16] Quoc L., and Tomas M., "Distributed representations of sentences and documents", *International Conference on Machine Learning*, pp. 1188-1196, 2014.

[17] Rajesh P., Devasher M., and Vivek Kumar S., "Analytical mapping of opinion mining and sentiment analysis research during 2000–2015", *Information Processing & Management*, vol.53, no.1, pp. 122-150, 2017.

[18] Salima M., et al., "Sentiment analysis of Tunisian dialects: Linguistic resources and experiments", *Proceedings of the Third Arabic Natural Language Processing Workshop*, pp.55-61, 2017.

[19] Samhaa E., and Ahmad A., "Open issues in the sentiment analysis of Arabic social media: A case study", *9th IEEE international conference on Innovations in information technology (iit)*, pp. 215-220, 2013.

[20] Samhaa E., et al., "Combining Lexical Features and a Supervised Learning Approach for Arabic Sentiment Analysis", *International Conference on Intelligent Text Processing and Computational Linguistics*, pp. 307-319, 2016.

[21] Tomas M., et al., "Distributed representations of words and phrases and their compositionality", *Advances in neural information processing system*, pp. 3111-3119, 2013.